\documentclass[12pt,english]{paper}
\usepackage[T1]{fontenc}
\usepackage[latin9]{inputenc}
\usepackage{float}
\usepackage{amsthm}
\usepackage{amsmath}
\usepackage{esint}

\makeatletter

\providecommand{\tabularnewline}{\\}

\numberwithin{equation}{section}
\numberwithin{figure}{section}
\newcommand{\lyxaddress}[1]{
\par {\raggedright #1
\vspace{1.4em}
\noindent\par}
}

\makeatother

\usepackage{babel}
\begin{document}

\title{Quantum Cybernetics and Complex Quantum Systems Science - A Quantum
Connectionist Exploration}

\author{Carlos Pedro Gonçalves}

\institution{Instituto Superior de Ciências Sociais e Políticas (ISCSP) - University
of Lisbon}

\maketitle

\lyxaddress{cgoncalves@iscsp.ulisboa.pt}
\begin{abstract}
Quantum cybernetics and its connections to complex quantum systems
science is addressed from the perspective of complex quantum computing
systems. In this way, the notion of an autonomous quantum computing
system is introduced in regards to quantum artificial intelligence,
and applied to quantum artificial neural networks, considered as autonomous
quantum computing systems, which leads to a quantum connectionist
framework within quantum cybernetics for complex quantum computing
systems. Several examples of quantum feedforward neural networks are
addressed in regards to Boolean functions' computation, multilayer
quantum computation dynamics, entanglement and quantum complementarity.
The examples provide a framework for a reflection on the role of quantum
artificial neural networks as a general framework for addressing complex
quantum systems that perform network-based quantum computation, possible
consequences are drawn regarding quantum technologies, as well as
fundamental research in complex quantum systems science and quantum
biology.\end{abstract}

\section{Introduction}

The current work is aimed at the expansion of quantum cybernetics
and complex quantum systems science by addressing directly two central
threads within quantum artificial intelligence research, namely: the
notion of an autonomous quantum computing system (AQCS) and a generalization
to the quantum setting of the connectionist framework for artificial
intelligence. Addressing both the computational aspects as well as
the implications for complex quantum systems science and for quantum
technologies.

In section 2., a general background on quantum cybernetics is provided,
including its connection with quantum computation and complex quantum
systems science. In section 3., the notion of an autonomous quantum
computing system (AQCS), in connection to quantum artificial intelligence,
is introduced. In section 4., quantum artificial neural networks (QuANNs)
theory is reviewed and QuANNs are addressed as AQCSs. In particular,
quantum feedforward neural networks are addressed and major points
such as the problem of computation of Boolean functions and neural
network computational complexity, multiple layers and the issue of
complementarity are addressed. In section 5., a reflection is presented,
addressing the main results and the connection between complex quantum
systems science, quantum connectionist approaches and quantum technologies.

\section{Quantum Cybernetics and Quantum Computation}

Cybernetics began as a post-World War II scientific movement gathering
researchers from different disciplinary backgrounds \cite{key-1}.
Norbert Wiener defined cybernetics as the scientific study of control
and communication in the animal and the machine \cite{key-2}. Intersecting
with general systems science, cybernetics grew as a transdisciplinary
research field dealing with issues like feedback, control and communication,
self-organization and autonomy in natural and artificial systems \cite{key-1}.
In artificial systems these issues are raised to the central research
problem of an autonomous computing system endowed with an artificial
intelligence that may allow it to select a program and adapt to different
conditions.

Up until the end of the 20th Century, cybernetics was strongly influenced,
in what regards autonomous computing systems, by Shannon's information
theory and by the works of Turing and of Von Neumann in computer science.
Turing distinguished himself particularly in the definition of an
automatic computing machine and a universal computing machine \cite{key-3},
as well as in artificial intelligence \cite{key-4}. Von Neumann proposed
an automaton theory and introduced the basis for the scientific field
of artificial life \cite{key-5,key-6}.

These works influenced different scientific disciplines and authors,
one in particular, Hugh Everett III, in his PhD thesis on quantum
mechanics \cite{key-7}, was the first to introduce cybernetics notions
to address the foundations of quantum theory, namely, by using information
theory's notions to address quantum theory and a notion of observer
defined, in Everett's proposal, as an automaton with a memory register
capable of interacting with the quantum system and becoming entangled
with it \cite{key-7,key-8}.

Despite this early work, quantum cybernetics, as a subfield within
cybernetics, is mainly the result of the advancement of research in:
quantum computation theory and quantum information theory; complex
quantum systems science and quantum game theory, during the 1970s,
1980s and 1990s.

The roots of quantum computation and quantum information theory can
be traced back to the work of Everett himself in his application of
information theory to the framework of quantum mechanics and his proposal
of a quantum automaton model to deal with the observer and the observation
act. Other roots can be found in several works during the 1970s and
1980s, namely, in the 1970s one can find, for instance, the work of
Weisner on quantum coding, in the research field of quantum cryptography,
which also led to quantum game theory \cite{key-9,key-10,key-11},
and Holevo's work on quantum information \cite{key-12}, as well as
other authors like Feynman, Beniof and Deutsch \cite{key-13,key-14,key-15},
who moved forward the research field of quantum computation. Deutsch,
in particular, introduced the notion of a universal quantum computing
machine based upon a quantum extension of Turing's machine, the research
field of quantum computation advanced considerably in terms of potential
practical applications with Shor's factoring algorithm and Grover's
search algorithm \cite{key-16}.

Today quantum computation theory is a growing multidisciplinary field
that intercrosses quantum theory with computer science and the complexity
sciences (in particular complex quantum systems science). The intersection
of quantum computation and quantum game theory, on the other hand,
opens the way for research on autonomous artificial systems, incorporating
quantum artificial intelligence, as well as for research in the growing
field of quantum biology%
\footnote{University of Surrey's Institute of Advanced Studies (IAS), hosted
a conference on quantum biology, jointly organized by IAS, the Biotechnology
and Biological Sciences Research Council (BBSRC) and MILES (Models
and Mathematics in Life and Social Sciences). The following link contains
more information on the conference including the presentations: http://www.ias.surrey.ac.uk/workshops/quantumbiology/.%
} \cite{key-17,key-18,key-19,key-20}. Quantum cybernetics deepens
the link between complex quantum systems science and (quantum) computer
science, with strong connections to the work developed on the foundations
of quantum theory and the role of information in quantum mechanics%
\footnote{An early work addressing these issues is Seth Lloyd's \cite{key-21}.%
}.

The recent creation of Google and NASA's {}``Quantum Artificial Intelligence
Laboratory''%
\footnote{http://www.nas.nasa.gov/quantum/%
}, in collaboration with D-Wave Systems%
\footnote{Some polemic has been associated with the D-Wave quantum computer,
in regards to whether or not it truly is a quantum computer. A review
of the arguments on each side would go beyond the scope of the present
work. It's collaboration with NASA, Google and Lockheed Martin, in
the development of quantum technologies, is sufficient to exemplify
investment in research and development on quantum technologies.%
} and Los Alamos' quantum internet, along with the research currently
being developed on quantum artificial intelligence, quantum cybernetics
can be considered to be gaining momentum in terms of the relation
between academic research and possible near-future practical applications.

Applications of quantum adaptive computation and quantum optimization
also extends to both quantum biology research and quantum game theory,
in the case of the later as an adaptive computing framework within
interdisciplinary fields such as econophysics, with proven effectiveness
in dealing, for instance, with financial risk modeling \cite{key-23,key-24}.

The current work addresses another direction within quantum cybernetics,
which is the extension to the quantum setting of the connectionist
approach to artificial intelligence, regarding the artificial neural
networks' research \cite{key-25,key-26,key-27,key-28,key-29}. In
the current article, three main points of cybernetics are, thus, addressed
and interlinked in regards to the quantum setting: the notion of autonomy
in complex computing systems; quantum artificial intelligence and
quantum artificial neural networks.

\section{Quantum Computation and Quantum Artificial Intelligence}

The main insight behind quantum computation regards the representation
of quantum information, while the most basic unit of classical information
can be expressed in terms of a binary digit called {}``bit'', which
is either zero or one, the most basic unit of quantum information
can be expressed in terms of a quantum binary digit or {}``qubit''.
The major difference is that the qubit corresponds to a superposition
of the classical information states. In quantum computation theory,
the qubit is addressed, formally, as a vector of a two-dimensional
Hilbert space $\mathcal{H}_{2}$ with computational basis $\left\{ \left|0\right\rangle ,\left|1\right\rangle \right\} $,
a qubit, using Dirac's bra-ket notation for vectors, is then defined
as a normalized ket vector:

\begin{equation}
\left|\psi\right\rangle =\psi(0)\left|0\right\rangle +\psi(1)\left|1\right\rangle 
\end{equation}
where the weights correspond to complex amplitudes satisfying the
normalization condition $\left|\psi(0)\right|^{2}+\left|\psi(1)\right|^{2}=1$.

Quantum computations on a single qubit can, thus, be addressed in
terms of the group U(2), the unitary group in $2$ dimensions, whose
elements correspond to quantum gates, using the quantum circuit terminology
\cite{key-16}. If we assume the following matrix representation of
the computational basis:

\begin{equation}
\left|0\right\rangle \equiv\left(\begin{array}{c}
1\\
0
\end{array}\right),\:\left|1\right\rangle \equiv\left(\begin{array}{c}
0\\
1
\end{array}\right)
\end{equation}
the general unitary operator can be shown to have the following matrix
representation:

\begin{equation}
\hat{U}_{\phi_{0},u,v}=e^{i\phi_{0}}\left(\begin{array}{cc}
u & v\\
-v^{*} & u^{*}
\end{array}\right)
\end{equation}
with the normalization condition $|u|^{2}+|v|^{2}=1$.

Thus, we have a global phase transformation $e^{i\phi_{0}}$ (an element
of the U(1) group) and a unitary $2\times2$ matrix with unit determinant
(an element of the $\textrm{SU}(2)$ group). The element of $\textrm{SU}(2)$
can be expressed as:

\begin{equation}
\left(\begin{array}{cc}
u & v\\
-v^{*} & u^{*}
\end{array}\right)=\left(\begin{array}{cc}
e^{i\phi_{1}}\cos\phi_{3} & e^{i\phi_{2}}\sin\phi_{3}\\
-e^{-i\phi_{2}}\sin\phi_{3} & e^{-i\phi_{1}}\cos\phi_{3}
\end{array}\right)
\end{equation}
so that, introducing the tuple $\boldsymbol{\phi}\equiv\left(\phi_{0},\phi_{1},\phi_{2},\phi_{3}\right)$,
we can rewrite, for the U(2) matrix:

\begin{equation}
\hat{U}_{\boldsymbol{\phi}}=e^{i\phi_{0}}\left(\begin{array}{cc}
e^{i\phi_{1}}\cos\phi_{3} & e^{i\phi_{2}}\sin\phi_{3}\\
-e^{-i\phi_{2}}\sin\phi_{3} & e^{-i\phi_{1}}\cos\phi_{3}
\end{array}\right)
\end{equation}

In the matrix representation, different operators $\hat{U}_{\boldsymbol{\phi}}$
can be addressed, with respect to the parameter range, under the notion
of matrix identity, indeed, working from the notion of matrix identity,
the $\textrm{U}(1)$ component is such that $e^{i(\phi_{0}+2\pi)}=e^{i\phi_{0}}$,
so that any two quantum gates differing in the $\textrm{U}(1)$ component
by a phase of $2\pi$ coincide, thus, the angle $\phi_{0}$ can be
defined in the interval $\left[0,2\pi\right]$, on the other hand,
still working with matrix identity, for a given $\phi_{0}$, the phases
for each entry also share this periodicity so that $\phi_{1}$ and
$\phi_{2}$ return the corresponding entry to the same element with
a $2\pi$ periodicity, leading to the interval $\left[0,2\pi\right]$
for both of these angles as well. Now considering the three angles
fixed $\phi_{0},\phi_{1},\phi_{2}$, we are left with the problem
of addressing a parametrization taking into account the trigonometric
functions, there are two trigonometric functions appearing in the
matrix entries: $\sin\phi_{3}$ and $\cos\phi_{3}$.

The range for the angle $\phi_{3}$ must be assumed, in this case,
between $0$ and $\pi/2$ for the reason that for $\phi_{3}>\pi/2$,
the element of U(2) can be written in terms of a matrix with an angle
between $0$ and $\pi/2$ with a different phase, which means that
the entire system of quantum gates is covered by $\phi_{3}$ taken
in the interval $\left[0,\pi/2\right]$. As way to make all of the
angles range in the same interval $\left[0,2\pi\right]$, we have
to rewrite the matrix as:
\begin{equation}
\hat{U}_{\boldsymbol{\phi}}=e^{i\phi_{0}}\left(\begin{array}{cc}
e^{i\phi_{1}}\cos\frac{\phi_{3}}{4} & e^{i\phi_{2}}\sin\frac{\phi_{3}}{4}\\
-e^{-i\phi_{2}}\sin\frac{\phi_{3}}{4} & e^{-i\phi_{1}}\cos\frac{\phi_{3}}{4}
\end{array}\right)
\end{equation}
assuming, now, also the range $\phi_{3}\in\left[0,2\pi\right]$.

This is equivalent to the polar representation of a 3-sphere, so that
$e^{i\phi_{1}}\cos\phi_{3}/4$ and $e^{i\phi_{2}}\sin\phi_{3}/4$,
with $\phi_{1},\phi_{3}\in\left[0,2\pi\right]$ correspond to a reparametrization
of the usual spherical coordinates: usually, the angle $\phi_{3}$
for the trigonometric functions is assumed as ranging in $0$ to $\pi$
and the division is set by $2$ instead of $4$, however, it is useful
to address the quantum decision dynamics of an AQCS, by letting all
angles range from $0$ to $2\pi$, which explains the division by
$4$, this due to a simplification of the Hamiltonian formulation
for the U(2) group wave function, a point to which we will return
further on.

As it stands, from the above representation, it follows that quantum
computation can be addressed in terms of quantum gates acting on individual
qubits, such that, in the quantum circuit model representation, information
travels through the {}``wires'' and is transformed by the {}``gates''
\cite{key-16}. This type of representation corresponds to a model
of a programmable quantum computing machine, with the quantum algorithm
defined in terms of the quantum unitary transformations (quantum gates).

However, if we wish to address quantum artificial intelligence (quantum
AI) we need to allow for autonomous quantum computation defining an
AQCS, which means that the quantum computing system must be capable
of adapting to different environmental conditions and respond accordingly,
thus, given an input qubit $\left|\psi_{in}\right\rangle $ we can
consider that for different states of an environment $\left|f\right\rangle $
the quantum AI must be able to evaluate this environment responding
with a different quantum program synthesized in the form of a quantum
gate applied to $\left|\psi_{in}\right\rangle $, in this sense, we
obtain a quantum conditional response to the environment in the form
of a conditional quantum computation with the alternative quantum
computation histories $\left|f\right\rangle \otimes\hat{U}_{f}\left|\psi_{in}\right\rangle $.
A specific case is the one where each alternative quantum computation
is performed, with a given amplitude:
\begin{equation}
\left|\Psi\right\rangle =\int d^{4}\boldsymbol{\phi}\Psi(\boldsymbol{\phi})\left|\boldsymbol{\phi}\right\rangle \otimes\hat{U}_{\boldsymbol{\phi}}\left|\psi_{in}\right\rangle 
\end{equation}
where $\Psi(\boldsymbol{\phi})$ is a quantum amplitude over the U(2)
parameter space, thus, for the above parameterization, it is only
different from zero for $\phi_{k}\in[0,2\pi]$, with $k=0,1,2,3$,
thus covering each of the alternative U(2) elements only once, under
the above parametrization of U(2). The ket $\left|\Psi\right\rangle $
is an expression of a quantum AI adapting to a quantum environment
which takes the form of an interaction that takes place coupling to
the U(2) transformation. Instead of a single quantum program, the
quantum program depends upon this interaction, in this sense, we can
speak of a quantum cognition, since the quantum AI responds to the
different envrionmental states with a different quantum computation.
We are dealing with a form of quantum hypercomputation, in the sense
that the system (environment+AQCS) leads to a (quantum) superposition
of the alternative quantum computations on the qubit.

One may also introduce a Hamiltonian unitary evolution for the environment,
which assumes that the U(2) quantum computation is related to a quantum
fluctuating environment that is described by a Hamiltonian description
of the U(2) group, using the parameter space, under the above parametrization,
for the definition, which leads to the change of (3.7) to the following
structure:
\begin{equation}
\left|\Psi\right\rangle =\int d^{4}\boldsymbol{\phi}e^{-i\hat{H}t}\Psi(\boldsymbol{\phi})\left|\boldsymbol{\phi}\right\rangle \otimes\hat{U}_{\boldsymbol{\phi}}\left|\psi_{in}\right\rangle 
\end{equation}

Due to the parametrization of the operators $\hat{U}_{\boldsymbol{\phi}}$,
where $\boldsymbol{\phi}$ is a four-tuple $\left(\phi_{0},\phi_{1},\phi_{2},\phi_{3}\right)$,
with each angle $\phi_{k}$ between $0$ and $2\pi$, one possible
choice of Hamiltonian may consider addressing the Hamiltonian in terms
of a box potential with the shape of a tesseract with each side measuring
$2\pi$ as the basic structure for the box (a four-dimensional box),
however, given the periodic nature of the quantum gates in U(2), for
the U(2) parametrization chosen, it follows that we should assume
periodic boundary conditions for the tesseract leading to a hypertorus
(in this case, a 4-torus). Considering the free Hamiltonian yields
the U(2) connection expressed in (3.8), so that we have a free U(2)
Schrödinger equation, under the previous parametrization, with periodic
boundary conditions on the hypertorus:
\begin{equation}
-\nabla^{2}\Psi=\lambda\Psi
\end{equation}

Separating the wave functions:
\begin{equation}
\Psi\left(\phi_{0},\phi_{1},\phi_{2},\phi_{3}\right)=\Psi\left(\phi_{0}\right)\Psi\left(\phi_{1}\right)\Psi\left(\phi_{2}\right)\Psi\left(\phi_{3}\right)
\end{equation}
we can solve the four free Schrödinger equations with the general
form:

\begin{equation}
-\frac{d^{2}}{d\phi_{k}}\Psi\left(\phi_{k}\right)=\lambda_{k}\Psi\left(\phi_{k}\right)
\end{equation}
assuming the periodic condition $\Psi\left(\phi_{k}+2\pi\right)=\Psi\left(\phi_{k}\right)$,
the solution to this last equation is:

\begin{equation}
\Psi_{n}\left(\phi_{k}\right)=\frac{1}{\sqrt{2\pi}}e^{in\phi_{k}}
\end{equation}
with the momentum eigenvalues:

\begin{equation}
n=0,\pm1,\pm2,...
\end{equation}
Letting $\mathbf{n}\equiv\left(n_{0},n_{1},n_{2},n_{3}\right)$, the
energy eigenvalues are:

\begin{equation}
\lambda_{\mathbf{n}}=n_{0}^{2}+n_{1}^{2}+n_{2}^{2}+n_{3}^{2}
\end{equation}
So that we have the final eigenfunctions: 
\begin{equation}
\Psi_{\mathbf{n}}\left(\boldsymbol{\phi}\right)=\left\langle \boldsymbol{\phi}|\mathbf{n}\right\rangle =\prod_{k=0}^{3}\frac{1}{\sqrt{2\pi}}e^{in_{k}\phi_{k}}=\frac{1}{4\pi^{2}}\exp\left(i\sum_{k=0}^{3}n_{k}\phi_{k}\right)
\end{equation}
The reason for the parametrization chosen is now made clear, in the
sense that the Hamiltonian condition and the wave function simplify
greatly leading to a free Schrödinger U(2) equation being expressed
as a free Schrödinger equation on a hypertorus (equation (3.9)).

For (3.8) we can now write:
\begin{equation}
\left|\Psi(t)\right\rangle =\int d^{4}\boldsymbol{\phi}\Psi(\boldsymbol{\phi},t)\left|\boldsymbol{\phi}\right\rangle \otimes\hat{U}_{\boldsymbol{\phi}}\left|\psi_{in}\right\rangle 
\end{equation}
where $\Psi(\boldsymbol{\phi},t)$ is the wave packet given by: 
\begin{equation}
\Psi(\boldsymbol{\phi},t)=\sum_{\mathbf{n}}A_{\mathbf{n}}e^{-i\lambda_{\mathbf{n}}t}\Psi_{\mathbf{n}}\left(\boldsymbol{\phi}\right)
\end{equation}
with the usual normalization condition assumed:
\begin{equation}
\sum_{\mathbf{n}}\left|A_{\mathbf{n}}\right|^{2}=1
\end{equation}
Replacing (3.15) in (3.17) we get the following solution for the wave
packet: 
\begin{equation}
\Psi(\boldsymbol{\phi},t)=\sum_{\mathbf{n}}A_{\mathbf{n}}\frac{1}{4\pi^{2}}\exp\left[i\left(\sum_{k=0}^{3}n_{k}\phi_{k}-\lambda_{\mathbf{n}}t\right)\right]
\end{equation}

Taking into account this framework, we are now ready to introduce
a quantum version of artificial neural networks based upon the AQCS
model. We will deal, in particular, with basic neural connections
and QuANNs.

\section{Quantum Artificial Neural Networks as Autonomous Quantum Computing
Systems}

The connectionist paradigm of computation is a strong example of the
intersection between computer science and the cybernetics' paradigmatic
base. In 1943, McCulloch and Pitts wrote an article entitled \emph{A
Logical Calculus of the Ideas Immanent in Nervous Activity} \cite{key-27},
which introduced another computing framework based in the function
of neurons, marking the birth of artificial neural networks, now a
classic of early cybernetics \cite{key-28}. McCulloch and Pitts'
proposal was the subject of intense debate within what later came
to be the connectionist-based paradigm of computation \cite{key-1,key-28,key-29}.

Artificial neural networks (ANNs), while largely based upon a simplification
of the human neural structure, form a basic setting for research on
network-based computation resulting from the agencing of a large number
of elementary computing units called neurons \cite{key-28,key-29}.
ANNs constitute a good mathematical framework for addressing computing
network systems as complex computing systems, allowing one to study
the relation between the network's architecture and its function.

Information in ANNs has a specific interpretation as well: while in
Turing's computational framework, information was addressed in terms
of symbols placed in a machine, in ANNs' computational framework,
information is associated with a pattern of activity, the system (the
network) is the entity, the system and the pattern of activity are,
thus, primitive and fundamental, information being the result of the
system's activity and not the {}``thing'' to be printed on a tape
(as takes place in Turing's machine), even the memory register can
be addressed in terms of the network's recollecting activity \cite{key-28,key-29}.
In this sense, ANNs are closer to systems science and to a biological
paradigm \cite{key-28,key-29}, in some way incorporating a bionics%
\footnote{Notion introduced by Jack E. Steele to refer to biologically-based
design of artificial and intelligent systems. Bionics was defended
by von Foerster, who greatly influenced the complexity sciences \cite{key-30},
mainly in what regards biology-based computation approaches. In this
sense, ANNs are already concurrent with bionics, as biology-inspired
directions within the paradigmatic base of cybernetics, that go beyond
a mechanistic approach \cite{key-1,key-28}.%
} base of research.

Quantum artificial neural networks (QuANNs) constitute the quantum
extension of ANNs. QuANNs are complex quantum computing systems and
allow one to harness the computing power of {}``networked'' quantum
computation.

Research on the role of quantum effects in biological neurodynamics
has been a topic of interdisciplinary research within neuroscience
and physics, with authors like Eccles and Beck, Penrose and Hameroff
that defend the hypothesis of quantum neurodynamics in the explanation
of consciousness, while other authors, like Tegmark have laid arguments
against that hypothesis \cite{key-32,key-33,key-34}, Tegmark's argument
relies on environmental decoherence associated to an interaction with
the environment that effectively destroys locally the off-diagonal
terms in a subsystem's density matrix, which implies a local entropy
increase. However, for this decoherence to take place the system+environment
have to become entangled, which means that we actually have macro-coherence
with subsystem decoherence.

We will see, in the present section, that entanglement can take place
in a neural network itself so that local neuron-level decoherence
is almost inevitable, however, one cannot state that quantum effects
are not present, parallel quantum processing in large networks tends
to produce entangled network states, in which the local node descripion,
or even several nodes' description shows local decoherence.

However, QuANNs are not necessarily restricted to research on quantum
neurodynamics, or even dependent upon the hypothesis of biological
quantum neurodynamics, rather, QuANNs are a model of network-based
quantum computation, constituting a framework for addressing parallel
quantum computation which can have multiple applications when one
deals with complex quantum systems that involve interacting components.
QuANNs may also come to play a role in future quantum technologies,
points to which we shall return in the last section of the present
work.

The connection between quantum theory and artificial neural network
theory has been developed, since the 1990s, in connection to quantum
computation, in particular, linked to quantum associative memory,
parallel processing and schemes of extension of ANNs to the quantum
setting \cite{key-35,key-36,key-37,key-38,key-39,key-40}.

Considering an artificial neuron as a computing unit with two states
0 (non-firing) and 1 (firing), which, in itself a far simpler model
than the biological neuron, allows for network-based classical computation,
exploring common aspects with biological neural networks. The quantum
extension of McCulloch and Pitts proposal leads one to address the
artificial neuron's two states as the basis states of the qubit, which
means that the artificial neuron can be in a superposition of the
non-firing state, represented by the \emph{ket} vector $\left|0\right\rangle $,
and the firing state, represented by the \emph{ket} vector $\left|1\right\rangle $.

Extending the quantum circuit approach to quantum computation, in
the parallel processing framework, means that we assign qubits to
neurons and quantum gates to the synaptic connections, the quantum
gates for the synaptic connections express the interaction between
the neurons in the form of a quantum computation, which means that
the synaptic connection is more complex than in classical ANNs, in
the sense that it plays an active role in allowing for network-based
quantum computation, including the possibility of adaptive quantum
computation.

To explore the range of possibilities we address, in the current work,
quantum feedforward neural networks. In its simplest form the (classical)
feedforward neural network consists of two layers of neurons: a first
layer of input neurons and a second layer of output neurons \cite{key-29}.
The neurons of the output layer receive synaptic signals from the
input layer. A more complex model includes multiple layers, allowing
for more complex computations \cite{key-29}. Quantum feedforward
neural networks can be used to compute all classical Boolean functions
in a more efficient way than classical feedforward neural networks,
as we now show.

\subsection{Quantum feedforward neural networks and Boolean function computation}

We will begin by addressing, in this section, examples of application
of quantum feedforward neural networks, in the examples we always
start with the network such that each neuron is in the non-firing
state $\left|0\right\rangle $, the quantum computation proceeds then
as a form of quantum learning dynamics through quantum parallel processing.
In the examples we will consider first the computation of Boolean
functions of the form $\left\{ 0,1\right\} ^{m}\rightarrow\left\{ 0,1\right\} $,
and afterwards generalize to the Boolean functions $\left\{ 0,1\right\} ^{m}\rightarrow\left\{ 0,1\right\} ^{n}$.
There are, for general $m$, $2^{2^{m}}$ such Boolean functions,
considering $m=1$, as a first illustrative example, we get the four
basic Boolean functions:

\begin{table}[H]
\begin{centering}
\begin{tabular}{|c|c|c|}
\hline 
$g(s)$ & $s=0$ & $s=1$\tabularnewline
\hline 
\hline 
$g(s)=0$ & $0$ & $0$\tabularnewline
\hline 
$g(s)=s$ & $0$ & $1$\tabularnewline
\hline 
$g(s)=1-s$ & $1$ & $0$\tabularnewline
\hline 
$g(s)=1$ & $1$ & $1$\tabularnewline
\hline 
\end{tabular}
\par\end{centering}

\caption{Boolean functions $\left\{ 0,1\right\} \rightarrow\left\{ 0,1\right\} $.}
\end{table}

The QuANN that is capable of computing these Boolean functions is
a two-layered quantum feedforward network:
\begin{equation}
N_{1}\longrightarrow N_{2}
\end{equation}
where $N_{1}$ is the input neuron and $N_{2}$ is the output neuron.
The initial state of the two neurons is $\left|00\right\rangle $
(both are non-firing), following the previous section's approach,
if we deal with the above network as an AQCS, the ket vector for the
network's quantum computation histories' ket is given by: 
\begin{equation}
\left|\Psi\right\rangle =\int d^{4}\boldsymbol{\phi}\Psi(\boldsymbol{\phi})\left|\boldsymbol{\phi}\right\rangle \otimes\hat{\mathcal{N}}_{\boldsymbol{\phi}}\left|00\right\rangle 
\end{equation}
where $\hat{\mathcal{N}}_{\boldsymbol{\phi}}$ is an operator on the
Hilbert space $\mathcal{H}_{2}\otimes\mathcal{H}_{2}$, corresponding
to a single quantum computation history for the quantum network, and
$\Psi(\boldsymbol{\phi})$ is the quantum amplitude associated with
each alternative history, in this case taken as a wave packet of the
form:
\begin{equation}
\Psi(\boldsymbol{\phi})=\sum_{\mathbf{n}}A_{\mathbf{n}}\Psi_{\mathbf{n}}\left(\boldsymbol{\phi}\right)
\end{equation}
such that the Hamiltonian condition, represented by the Schrödinger
equation (3.9), becomes a Hamiltonian restriction for the network's
quantum computation. The structure of $\hat{\mathcal{N}}_{\boldsymbol{\phi}}$
is such that the input neuron becomes the input qubit and the output
neuron plays the role of a generalized Everett automaton that interacts
with the input neuron through the synaptic connection, more complex
quantum computation types will be considered in the next section.
Given this structure, in regards to the Boolean function computation,
$\hat{\mathcal{N}}_{\boldsymbol{\phi}}$ takes the form: 
\begin{equation}
\hat{\mathcal{N}}_{\boldsymbol{\phi}}=\hat{S}_{g}\hat{U}_{\boldsymbol{\phi}}\otimes\hat{1}
\end{equation}
thus, $\hat{\mathcal{N}}_{\boldsymbol{\phi}}$ is such that a U(2)
transformation is first applied to the input neuron leaving the output
neuron unchanged (operator $\hat{U}_{\boldsymbol{\phi}}\otimes\hat{1}$)
and, then, $\hat{S}_{g}$ is applied representing the neural network's
internal networked processing, where $\hat{S}_{g}$ is a unitary operator
on the Hilbert space $\mathcal{H}_{2}\otimes\mathcal{H}_{2}$, whose
structure depends upon the Boolean function under computation, in
the neural network model, this operator expresses the structure of
the synaptic connection which determines the interaction pattern between
the two neurons. The following table shows the general structure for
the synaptic connection quantum gates $\hat{S}_{g}$ and the corresponding
alternative output kets associated with each quantum computation histories:

\begin{table}[H]
\begin{centering}
\begin{tabular}{|c|c|c|}
\hline 
$g(s)$ & $\hat{S}_{g}$ & Output Kets\tabularnewline
\hline 
\hline 
$g(s)=0$ & $\hat{1}\otimes\hat{1}$ & $\left|\psi_{\boldsymbol{\phi}}0\right\rangle $\tabularnewline
\hline 
$g(s)=s$ & $\left|0\right\rangle \left\langle 0\right|\otimes\hat{1}+\left|1\right\rangle \left\langle 1\right|\otimes\hat{U}_{NOT}$ & $\psi_{\boldsymbol{\phi}}(0)\left|\boldsymbol{\phi},00\right\rangle +\psi_{\boldsymbol{\phi_{1}}}(1)\left|\boldsymbol{\phi},11\right\rangle $\tabularnewline
\hline 
$g(s)=1-s$ & $\left|0\right\rangle \left\langle 0\right|\otimes\hat{U}_{NOT}+\left|1\right\rangle \left\langle 1\right|\otimes\hat{1}$ & $\psi_{\boldsymbol{\phi_{1}}}(0)\left|\boldsymbol{\phi},01\right\rangle +\psi_{\boldsymbol{\phi_{1}}}(1)\left|\boldsymbol{\phi},10\right\rangle $\tabularnewline
\hline 
$g(s)=1$ & $\hat{1}\otimes\hat{U}_{NOT}$ & $\left|\psi_{\boldsymbol{\phi}_{1}},1\right\rangle $\tabularnewline
\hline 
\end{tabular}
\par\end{centering}

\caption{Output kets for the neural network computing the Boolean functions
$\left\{ 0,1\right\} \rightarrow\left\{ 0,1\right\} $.}
\end{table}

In the first case, $\hat{S}_{g}$ has the structure: 
\begin{equation}
\hat{S}_{g}=\sum_{s=0}^{1}\left(\left|s\right\rangle \left\langle s\right|\otimes\hat{1}\right)=\hat{1}\otimes\hat{1}
\end{equation}
therefore, the synaptic connection is neutral, in the sense that the
activation pattern of the input neuron in nothing affects the activation
pattern of the output neuron, thus, for an initial non-firing configuration
for the two neurons $\left|00\right\rangle $, the quantum computation
histories are given by:
\begin{equation}
\hat{\mathcal{N}}_{\boldsymbol{\phi}}=\left(\hat{1}\otimes\hat{1}\right)\left(\hat{U}_{\boldsymbol{\phi}}\otimes\hat{1}\right)=\hat{U}_{\boldsymbol{\phi}}\otimes\hat{1}
\end{equation}
which means that the input neuron's state undergoes the unitary transformation
$\hat{U}_{\boldsymbol{\phi}}$, but it does not affect the output
neuron (neutral synaptic connection), thus, the output neuron remains
non-firing, independently of the input neuron, so that replacing (4.6)
in (4.2) we obtain:
\begin{equation}
\left|\Psi\right\rangle =\int d^{4}\boldsymbol{\phi}\Psi(\boldsymbol{\phi})\left|\boldsymbol{\phi},\psi_{\boldsymbol{\phi}}0\right\rangle 
\end{equation}
where we took $\left|\boldsymbol{\phi},\psi_{\boldsymbol{\phi}}0\right\rangle =\left|\boldsymbol{\phi}\right\rangle \otimes\left|\psi_{\boldsymbol{\phi}}\right\rangle \otimes\left|0\right\rangle $,
with $\left|\psi_{\boldsymbol{\phi}}\right\rangle =\psi_{\boldsymbol{\phi}}(0)\left|0\right\rangle +\psi_{\boldsymbol{\phi}}(1)\left|1\right\rangle $,
for the amplitudes resulting from the U(2) transformation: $\psi_{\boldsymbol{\phi}}(0)=\exp\left[i\left(\phi_{0}+\phi_{1}\right)\right]\cos\left(\phi_{3}/4\right)$
and $\psi_{\boldsymbol{\phi}}(1)=-\exp\left[i\left(\phi_{0}-\phi_{2}\right)\right]\sin\left(\phi_{3}/4\right)$.

For the second and third cases, the output neuron functions similarly
to a quantum memory register, so that the system has an incorporated
Everett-like automaton structure. For the second case, $\hat{S}_{g}$
has the structure of a controlled-not gate%
\footnote{We write the controlled-not $\hat{U}_{CNOT(1)}$ with the subscript
$CNOT(1)$ instead of the usual notation $CNOT$, this is done because
we will work, below, with another controlled not operator that performs
the negation when the control qubit is $\left|0\right\rangle $ rather
than $\left|1\right\rangle $, this other operator will be denoted
as $\hat{U}_{CNOT(0)}$.%
}:
\begin{equation}
\hat{S}_{g}=\left|0\right\rangle \left\langle 0\right|\otimes\hat{1}+\left|1\right\rangle \left\langle 1\right|\otimes\hat{U}_{NOT}=\hat{U}_{CNOT(1)}
\end{equation}
where $\hat{U}_{NOT}$ is the unitary {}``NOT'' gate:
\begin{equation}
\hat{U}_{NOT}=\left|0\right\rangle \left\langle 1\right|+\left|1\right\rangle \left\langle 0\right|
\end{equation}

The synaptic connection, in this case, corresponds to a reinforcing
synaptic connection, thus, for an initial non-firing configuration
for the two neurons $\left|00\right\rangle $, the quantum computation
histories are given by:
\begin{equation}
\hat{\mathcal{N}}_{\boldsymbol{\phi}}=\hat{U}_{CNOT(1)}\hat{U}_{\boldsymbol{\phi}}\otimes\hat{1}
\end{equation}
which means that a quantum fluctation takes place at the input neuron
and then the synaptic interaction with the second neuron leads to
an entangled state, in which either both neurons fire or not.

It is important to notice that the entangled state shows a quantum
correlation, in the sense that the configuration is either the two
neurons fire synchronously without delay or not.

The synaptic connection acts in such a way as to allow an interaction
where the second neuron effectively performs a von Neumann measurement
upon the first neuron and, thus, acts in a similar way to an Everett
automaton's memory register dynamics.

For the third case, the situation is reversed, instead of a controlled-not
operator on $\left|1\right\rangle $, the controlled negation uses
$\left|0\right\rangle $ as the control:
\begin{equation}
\hat{S}_{g}=\left|0\right\rangle \left\langle 0\right|\otimes\hat{U}_{NOT}+\left|1\right\rangle \left\langle 1\right|\otimes\hat{1}=\hat{U}_{CNOT(0)}
\end{equation}
the synaptic connection, in this case, is inhibitory. As in the previous
case a quantum fluctuation takes place at the input neuron and the
synaptic interaction with the second neuron leads to an entangled
state, however, in this case, if one of the neuron is firing the other
is not firing and vice-versa. Replacing in (4.2), the second and third
cases lead, respectively, to the kets:
\begin{equation}
\left|\Psi\right\rangle =\int d^{4}\boldsymbol{\phi}\Psi(\boldsymbol{\phi})\left(\psi_{\boldsymbol{\phi}}(0)\left|\boldsymbol{\phi},00\right\rangle +\psi_{\boldsymbol{\phi}}(1)\left|\boldsymbol{\phi},11\right\rangle \right)
\end{equation}
\begin{equation}
\left|\Psi\right\rangle =\int d^{4}\boldsymbol{\phi}\Psi(\boldsymbol{\phi})\left(\psi_{\boldsymbol{\phi}}(0)\left|\boldsymbol{\phi},01\right\rangle +\psi_{\boldsymbol{\phi}}(1)\left|\boldsymbol{\phi},10\right\rangle \right)
\end{equation}

In the last case, the quantum computation histories are of the kind:
\begin{equation}
\hat{\mathcal{N}}_{\boldsymbol{\phi}}=\hat{1}\otimes\hat{U}_{NOT}\hat{U}_{\boldsymbol{\phi}}\otimes\hat{1}
\end{equation}
this means that while the input neuron is placed in a superposition,
the second neuron fires no matter what happens to the input neuron,
this can be considered as a global transformation of both neurons
that lead to a superposition state for the input neuron and to the
firing of the output neuron, the network's quantum computation histories'
ket is, in this case, given by:
\begin{equation}
\left|\Psi\right\rangle =\int d^{4}\boldsymbol{\phi}\Psi(\boldsymbol{\phi})\left|\boldsymbol{\phi},\psi_{\boldsymbol{\phi}}1\right\rangle 
\end{equation}

From the four cases, for the synaptic connection quantum gate, one
can see that its general form can be expressed as follows:
\begin{equation}
\hat{S}_{g}=\sum_{s=0}^{1}\left|s\right\rangle \left\langle s\right|\otimes\hat{U}_{g(s)}
\end{equation}
where the operator $\hat{U}_{g(s)}$ has the following structure:
\begin{equation}
\hat{U}_{g(s)=0}=\hat{U}_{0}\equiv\hat{1}
\end{equation}
\begin{equation}
\hat{U}_{g(s)=1}=\hat{U}_{1}\equiv\hat{U}_{NOT}
\end{equation}
thus if the Boolean function evaluates to $0$, then, the operator
$\hat{U}_{g(s)}$ coincides with the unit operator $\hat{1}$ on the
two-dimensional Hilbert space $\mathcal{H}_{2}$, since the initial
configuration of the network is always set so that the output neuron
is not firing, $\hat{U}_{g(s)=0}$ this means that the output neuron
does not fire when the input neuron exhibits the activity pattern
$\left|s\right\rangle $, on the other hand, when the Boolean function
evaluates to $1$, the operator $\hat{U}_{g(s)}$ coincides with the
NOT quantum gate $\hat{U}_{NOT}$, since the initial configuration
of the network is always set so that the output neuron is not initially
firing, this means that the output neuron fires when the input neuron
exhibits the activity pattern $\left|s\right\rangle $. This can be
generalized, in the sense that controlled negations can be used to
compute any Boolean function, as we now show.

Considering, first, the case of Boolean functions $\left\{ 0,1\right\} ^{m}\rightarrow\left\{ 0,1\right\} $,
the quantum feedforward neural network that is able to compute these
functions is comprised of $m$ input neurons and one output neuron,
without need of hidden layers, the network's quantum computation histories'
ket is given by: 
\begin{equation}
\left|\Psi\right\rangle =\int d^{4}\boldsymbol{\phi}_{m}...d^{4}\boldsymbol{\phi}_{1}\Psi(\boldsymbol{\phi}_{1},...,\boldsymbol{\phi}_{m})\left|\boldsymbol{\phi}_{1},...,\boldsymbol{\phi}_{m}\right\rangle \otimes\hat{\mathcal{N}}_{\boldsymbol{\phi}_{1},...,\boldsymbol{\phi}_{m}}\left|00...0;0\right\rangle 
\end{equation}
where we used, for the ket notation, the semicolon symbol {}``;''
to separate, in the ket, the input neurons' states from the output
neuron's state, and the operator $\hat{\mathcal{N}}_{\boldsymbol{\phi_{1},...,\phi_{m}}}$
is given by: 
\begin{equation}
\hat{\mathcal{N}}_{\boldsymbol{\phi_{1},...,\phi_{m}}}=\hat{S}_{g}\left(\bigotimes_{k=1}^{m}\hat{U}_{\boldsymbol{\phi}_{k}}\right)\otimes\hat{1}
\end{equation}
where $\hat{S}_{g}$ has the structure:
\begin{equation}
\hat{S}_{g}=\sum_{s_{1}...s_{m}}\left|s_{1}...s_{m}\right\rangle \left\langle s_{1}...s_{m}\right|\otimes\hat{U}_{g(s_{1}...s_{m})}
\end{equation}
such that if $g(s_{1}...s_{m})=0$, $\hat{U}_{g(s_{1}...s_{m})}=\hat{1}$
while if $g(s_{1}...s_{m})=1$, $\hat{U}_{g(s_{1}...s_{m})}=\hat{U}_{NOT}$.

In this case, the quantum neural networks grow with the number of
inputs, which reduces the structural complexity of the networks as
compared for instance to the Boolean function computation by a classical
feed-forward neural network with a hidden layer. While, for the classical
feedforward neural networks, the input layer grows with $m$, it is
possible to compute every Boolean function of the form $\left\{ 0,1\right\} ^{m}\rightarrow\left\{ 0,1\right\} $,
by a scheme that includes a hidden layer of neurons that grows as
$2^{m}$ \cite{key-29}. While the quantum feedforward neural network
does not need the hidden layer and grows in proportion to $m$. As
stressed in \cite{key-29} the classical case is a computationally
hard problem (NP-complete). The problem is reduced in complexity in
the quantum computation setting since the number of nodes does not
grow as $m+2^{m}$ ($m$ inputs plus $2^{m}$ hidden neurons), it
only grows as $m$ (the number of input neurons), without a hidden
layer. This result can be extended to the Boolean functions $\left\{ 0,1\right\} ^{m}\rightarrow\left\{ 0,1\right\} ^{n}$.

In this case, we use the following notation: if $g$ is a Boolean
function such that $g:\left\{ 0,1\right\} ^{m}\rightarrow\left\{ 0,1\right\} ^{n}$,
then, for an input string $s_{1}s_{2}...s_{m}$ if $g$ evaluates
to $g(s_{1}...s_{m})=s_{1}'s_{2}'...s_{n}'$ we define the local mappings
$g_{l}$, $l=1,2,...,n$ as: 
\begin{equation}
g_{l}(s_{1}...s_{m})=s_{l}'
\end{equation}
where $s_{l}'$ is the $l$-th symbol of the output string $s_{1}'s_{2}'...s_{n}'$.
With this notation in place, in order to compute the general Boolean
functions $\left\{ 0,1\right\} ^{m}\rightarrow\left\{ 0,1\right\} ^{n}$,
we need a quantum feedforward neural network with $m$ input neurons
and $n$ output neurons (once again, the network grows as $m+n$).
Assuming that the initial state of all neurons (input and output)
is non-firing, the quantum computation histories' ket is given by:

\begin{equation}
\begin{aligned}\left|\Psi\right\rangle =\int d^{4}\boldsymbol{\phi}_{m}...d^{4}\boldsymbol{\phi}_{1}\Psi(\boldsymbol{\phi}_{1},...,\boldsymbol{\phi}_{m})\left|\boldsymbol{\phi}_{1},...,\boldsymbol{\phi}_{m}\right\rangle \otimes\\
\otimes\hat{\mathcal{N}}_{\boldsymbol{\phi}_{1},...,\boldsymbol{\phi}_{m}}\left|00...0;00...0\right\rangle 
\end{aligned}
\end{equation}
where, once more, we used the semicolon symbol {}``$;$'' to separate,
in the ket, the input neurons' states from the output neurons' states.
As before, the computational histories $\hat{\mathcal{N}}_{\boldsymbol{\phi}_{1},...,\boldsymbol{\phi}_{m}}$
are defined as:

\begin{equation}
\hat{\mathcal{N}}_{\boldsymbol{\phi}_{1},...,\boldsymbol{\phi}_{m}}=\hat{S}_{g}\left(\bigotimes_{k=1}^{m}\hat{U}_{\boldsymbol{\phi}_{k}}\right)\otimes\hat{1}_{n}
\end{equation}
where $\hat{1}_{n}$ is the unit operator on the $2^{n}$ dimensional
Hilbert space for the output neurons' states given by the $n$ tensor
product of copies of $\mathcal{H}_{2}$. The operator $\hat{S}_{g}$,
in this case, has the structure:
\begin{equation}
\hat{S}_{g}=\sum_{s_{1}...s_{m}}\left|s_{1}...s_{m}\right\rangle \left\langle s_{1}...s_{m}\right|\otimes\hat{B}_{g(s_{1}...s_{m})}
\end{equation}
where $\hat{B}_{g(s_{1}...s_{m})}$ corresponds to the Boolean function
operator defined as: 
\begin{equation}
\hat{B}_{g(s_{1}...s_{m})}=\bigotimes_{l=1}^{n}\hat{U}_{g_{l}(s_{1}...s_{m})}
\end{equation}
where the local unitary operators $\hat{U}_{g_{l}(s_{1}...s_{m})}$
are defined such that $\hat{U}_{g_{l}(s_{1}...s_{m})}=\hat{1}$ when
$g_{l}(s_{1}...s_{m})=0$ and $\hat{U}_{g_{l}(s_{1}...s_{m})}=\hat{U}_{NOT}$
when $g_{l}(s_{1}...s_{m})=1$, in this way, the Boolean function
operator leads to a firing of each neuron or not, depending upon the
Boolean function's local configuration evaluating to either $0$ or
$1$, respectively. Thus, the interaction between the input and output
neurons, expressed by the operator $\hat{S}_{g}$, leads to the neural
computation of the Boolean functions, so that the network yields the
correct output through the pattern of activity of the output layer.
Only two layers are, once more, needed: an input layer of $m$ input
neurons and an ouput layer of $n$ output neurons, so that the network
grows in size as $m+n$.

While two layers allow for Boolean function computation, multiple
layers can be used for multiple Boolean function computations as well
as more complex neural network configurations, as we now show.

\subsection{Multiple layers and generalized quantum processing in quantum feedforward
neural networks}

Let us consider the quantum feedforward neural network with the following
architecture:
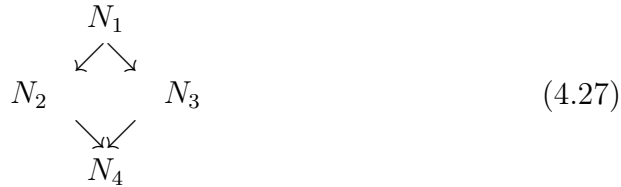
\begin{equation}
\begin{array}{ccc}
 & N_{1}\\
 & \swarrow\searrow\\
N_{2} &  & N_{3}\\
 & \searrow\swarrow\\
 & N_{4}
\end{array}
\end{equation}

The second layer allows for the computation of the Boolean functions
$g:\left\{ 0,1\right\} \rightarrow\left\{ 0,1\right\} ^{2}$, while
the third layer allows for the computation of the Boolean functions
$h:\left\{ 0,1\right\} ^{2}\rightarrow\left\{ 0,1\right\} $, using
the last section's framework, we separate each layer's state by a
semicolon, so that initially we have the network configuration where
each neuron is not firing, expressed by the ket: 
\begin{equation}
\left|0;00;0\right\rangle 
\end{equation}
The quantum histories for the quantum neural network are given by:
\begin{equation}
\mathcal{\mathcal{N}_{\boldsymbol{\phi}}}=\hat{S}_{h}\hat{S}_{g}\left(\hat{U}_{\boldsymbol{\phi}}\otimes\hat{1}_{3}\right)
\end{equation}
thus, as before, the unitary gate $\hat{U}_{\boldsymbol{\phi}}$ is
applied to the input neuron state, leaving the output neurons unchanged,
then, the first Boolean function is implemented, represented by the
operator $\hat{S}_{g}$ (which moves to the second layer) and the
second Boolean function is implemented, represented by the operator
$\hat{S}_{h}$, the operators are, respectively, given by:
\begin{equation}
\hat{S}_{g}=\left(\sum_{s}\left|s\right\rangle \left\langle s\right|\otimes\hat{B}_{g(s)}\right)\otimes\hat{1}
\end{equation}
\begin{equation}
\hat{S}_{h}=\hat{1}\otimes\left(\sum_{s,s'}\left|ss'\right\rangle \left\langle ss'\right|\otimes\hat{B}_{h(ss')}\right)
\end{equation}
with the Boolean operators given by:
\begin{equation}
\hat{B}_{g(s)}=\hat{U}_{g_{1}(s)}\otimes\hat{U}_{g_{2}(s)}
\end{equation}
\begin{equation}
\hat{B}_{h(ss')}=\hat{U}_{h(ss')}
\end{equation}
If, for instance, the functions are defined as:
\begin{equation}
g(0)=01,\: g(1)=10
\end{equation}
\begin{equation}
h(ss)=0,\: h(s1-s)=1
\end{equation}
The quantum computation histories' ket, resulting from the above equations,
is given by: 
\begin{equation}
\left|\Psi\right\rangle =\int d^{4}\boldsymbol{\phi}\Psi(\boldsymbol{\phi})\left(\psi_{\boldsymbol{\phi}}(0)\left|\boldsymbol{\phi},0;01;1\right\rangle +\psi_{\boldsymbol{\phi}}(1)\left|\boldsymbol{\phi},1;10;1\right\rangle \right)
\end{equation}

Therefore, in this case, the second neuron becomes entangled with
the input, through a reinforcing synaptic connection, while the third
neuron becomes entangled with the input neuron through a inhibitory
synaptic connection, which means that the two neurons' states in the
middle layer are no longer separable, they share a quantum correlation
with the input and, thus, with each other, such that the first two
layers corresponde to a fully entangled QuANN, the fourth neuron fires
if the second neuron fires and the third does not or if the second
neuron does not fire and the third fires, but it does not fire if
both the second and third neurons fire or if both the second and third
neurons do not fire, which is in accordance with the fact that $\hat{S}_{h}$
is computing the exclusive-or Boolean function at the third layer
of the neural network.

Since the second and third neurons share an inhibitory quantum correlation
it follows that the fourth neuron always fires. In this way, the fourth
neuron allows one to introduce a level of reflexivity regarding the
neural network's activity, in the sense that the fourth neuron fires
with the computation of a specific neural configuration in the middle
layer. It is, thus, possible to build {}``awareness'' neurons that
is: neurons that fire when certain patterns in the neural network
take place, and do not fire otherwise. Crossing levels one could even
introduce another link from the input neuron $N_{1}$ to $N_{4}$,
and make the computation conditional on the specific patterns of the
network up to that point.

While the efficient computation of Boolean functions is a central
point of the application of quantum feedforward neural networks, one
can go beyond the Boolean logical calculus and define more general
synaptic quantum gates.

Thus, for instance, for the same network architecture (4.27), if $\hat{U}_{h}$
is replaced by the following operator:
\begin{equation}
\hat{U}=\hat{U}_{H}\hat{U}_{h}
\end{equation}
where $\hat{U}_{H}$ is the quantum Haddamard transform:

\begin{equation}
\hat{U}_{H}=\frac{1}{\sqrt{2}}\left(\left|0\right\rangle \left\langle 0\right|-\left|1\right\rangle \left\langle 1\right|+\left|0\right\rangle \left\langle 1\right|+\left|1\right\rangle \left\langle 0\right|\right)
\end{equation}
and the Boolean function is replaced by:
\begin{equation}
h(00)=h(11)=0
\end{equation}
\begin{equation}
h(01)=0,\: h(10)=1
\end{equation}
then, for this new sequence of quantum computations, $\hat{\mathcal{N}}_{\boldsymbol{\phi}}$
leads to the following transformation: 
\begin{equation}
\hat{\mathcal{N}}_{\boldsymbol{\phi}}\left|0;00;0\right\rangle =\psi_{\boldsymbol{\phi}}(0)\left|0;01\right\rangle \otimes\left|+\right\rangle +\psi_{\boldsymbol{\phi}}(1)\left|1;10\right\rangle \otimes\left|-\right\rangle 
\end{equation}
with $\left|+\right\rangle =\hat{U}_{H}\left|0\right\rangle $ and
$\left|-\right\rangle =\hat{U}_{H}\left|1\right\rangle $, thus: 
\begin{equation}
\left|\pm\right\rangle =\frac{1}{\sqrt{2}}\left(\left|0\right\rangle \pm\left|1\right\rangle \right)
\end{equation}
The final ket, for the quantum computation histories, is no longer
(4.36) but the following:
\begin{equation}
\left|\Psi\right\rangle =\int d^{4}\boldsymbol{\phi}\Psi(\boldsymbol{\phi})\left(\psi_{\boldsymbol{\phi}}(0)\left|\boldsymbol{\phi},0;01\right\rangle \otimes\left|+\right\rangle +\psi_{\boldsymbol{\phi}}(1)\left|\boldsymbol{\phi},1;10\right\rangle \otimes\left|-\right\rangle \right)
\end{equation}
which means that, for each alternative branch, the last neuron is
always in a superposition between firing and non-firing. Even though
the last neuron is effectively entangled with the rest of the network,
the entanglement is for the alternative basis (4.42), in which the
neuron does not have a well defined firing pattern, showing rather
a superposition between two alternative firing patterns.

A quantum feedforward neural network can yield such superpositions
in the sense that it can generate a complementarity between firing
patterns, in the sense that if a neuron is in a well defined firing
activity, then, the other is in a superposition between firing and
not firing and vice-versa, this is a specifically quantum effect.
For classical ANNs this cannot take place. While (4.43) provides for
a more complex example, the simplest example can be obtained for a
two-layered quantum feedforward neural network with the architecture
(4.1). In the following structure for $\hat{\mathcal{N}}_{\boldsymbol{\phi}}$:
\begin{equation}
\hat{\mathcal{N}}_{\boldsymbol{\phi}}=\hat{S}_{12}\hat{U}_{\boldsymbol{\phi}}\otimes\hat{1}
\end{equation}
\begin{equation}
\hat{S}_{12}=\left|0\right\rangle \left\langle 0\right|\otimes\left(\hat{U}_{H}\hat{1}\right)+\left|1\right\rangle \left\langle 1\right|\otimes\left(\hat{U}_{H}\hat{U}_{NOT}\right)
\end{equation}
once again this leads to the result:
\begin{equation}
\left|\Psi\right\rangle =\int d^{4}\boldsymbol{\phi}\Psi(\boldsymbol{\phi})\left(\psi_{\boldsymbol{\phi}}(0)\left|\boldsymbol{\phi},0\right\rangle \otimes\left|+\right\rangle +\psi_{\boldsymbol{\phi}}(1)\left|\boldsymbol{\phi},1\right\rangle \otimes\left|-\right\rangle \right)
\end{equation}
therefore, in the branch where the input neuron fires, the ouput neuron
is in a superposition between firing and not firing described by $\left|+\right\rangle $
, similarly, in the branch where the input neuron does not fire, the
output neuron is in a superposition between firing and not firing
described by $\left|-\right\rangle $, for each alternative neural
computing history, the input neuron is entangled with the output neuron,
only it is not entangled with respect to the basis $\left\{ \left|0\right\rangle ,\left|1\right\rangle \right\} $,
but, rather, with respect to the basis $\left\{ \left|+\right\rangle ,\left|-\right\rangle \right\} $,
which means that the first neuron and the second neuron share a similarity
to the position and momentum in quantum mechanics, expressed by the
notion of quantum complementarity, if the input is in a well defined
firing pattern (either not firing or firing) the output no longer
has a well defined firing pattern (a superposition between firing
and not firing).

\section{Complex Quantum Systems and Quantum Connectionism}

In terms of technological development, so far, we have not yet reached
the point of building a QuANN, so that QuANNs largely constitute a
theoretical ground for exploring the properties of complex quantum
computing systems, and provide for a relevant line for the expansion
of quantum cybernetics in dialogue with complex quantum systems science.

The quantum connectionist framework of QuANNs can be understood as
a general framework for dealing with {}``networked'' quantum computation,
and not necessarily limited to the quantum neural processing conjecture.
In different fields of application of quantum theory one can find
the possible usefulness of incorporating QuANNs as conceptual tools
for addressing self-organization at the quantum level, for instance,
Zizzi, in her quantum computational approach to loop quantum gravity,
reached a development point that can be identified as an intersection
between spin networks and QuANNs \cite{key-41,key-42}. 

Considering QuANNs as complex quantum computing systems we can see,
from the examples of the quantum feedforward neural networks addressed
in the present work, that once we allow for an ANN to support networked
quantum computation through quantum superposition of neural firing
patterns and through quantum gates expressing synaptic connectivity
quantum dynamics, new properties start to emerge, that are not present
for classical ANNs: the system is capable of quantum hypercomputation,
it is possible to compute efficiently all the Boolean functions without
over-increasing the number of neurons, and new entangled states are
possible where the network may exhibit complementarity between entangled
neurons, such that if a neuron is firing the other is in a superposition
of neural states.

The examples, addressed in the current work, show how QuANNs may have
highly entangled states and can lead to a greater complexity in terms
of quantum computation. Three major points were introduced here that
are directly connected to quantum cybernetics, namely: the issue of
autonomy; the issue of hypercomputation and the issue of reflexivity
associated with quantum neural computation.

To go from a standard programmable quantum computer to a complex quantum
computing system, we introduced the notion of an autonomous quantum
computing system (AQCS), such that the system has an incorporated
artificial intelligence that allows it to choose from different alternative
quantum circuits in response to an interaction with a quantum environment
that couples to the U(2) gates, the interaction between system and
environment is, in this case, adaptive (that is, it triggers a quantum
computation) and leads to a simultaneous evaluation of each alternative
quantum computation history. In the case of QuANNs this framework
is complexified by the fact that the entire network's quantum computation
is responsive to the environment, namely the interaction with the
environment triggers a few neuron's quantum computation which in turn
triggers the synaptic interactions allowing for complex responses,
the entire network's activity is in this way responsive to the environment,
so that the alternative quantum computations $\mathcal{\mathcal{\hat{N}}}_{\boldsymbol{\phi_{1}},\boldsymbol{\phi_{2}},...,\boldsymbol{\phi_{m}}}$
that are implemented with corresponding amplitudes $\Psi(\boldsymbol{\phi}_{1},\boldsymbol{\phi}_{2},...,\boldsymbol{\phi}_{m})$
synthesize a level of reflexive response to the quantum environment.

In the case of the feedforward quantum neural networks the operators
$\mathcal{\mathcal{\hat{N}}}_{\boldsymbol{\phi_{1}},\boldsymbol{\phi_{2}},...,\boldsymbol{\phi_{m}}}$
lead to quantum computations that can cross the different layers of
the feedforward neural network, so that the whole network is computing
the initial neural stimulus. This is a form of hypercomputation in
the sense that alternative quantum computations can be performed with
different amplitudes throughout the neural network.

The network, thus, shows a level of distributed reflexivity in the
sense that the whole of the network reflects in a complex way the
initial neural stimulus that leads to the quantum fluctuation of the
input neurons. In this distributed reflexivity not only can the network
possess neurons that fire when they identify certain quantum neural
patterns (leading to a level of network-based reflexivity), but the
network itself may show a level of diversity in the neural patterns
exploring the quantum complementarity for different neurons, thus
exhibiting complex configurations that can only take place once we
go to the quantum computation framework.

The possibility to solve NP-hard problems of network-based computing,
with a smaller amount of computing resources is one of the points
that may support some researchers' conjecture of quantum biocomputation
in actual biological systems, which is already supported by empirical
evidence \cite{key-19,key-20}. In what regards neurobiology, this
entails the possibility of a form of quantum biological neural processing,
however, as previously stated, the community is divided on this last
issue. It is also important to stress that Damásio's work on neurobiology
showed that an organism's cognition cannot be reduced to the brain
activity, rather, the brain is an organ in an integrated system which
is the organism itself, Damásio showed that the whole body is involved
in producing consciousness \cite{key-43,key-44}, in terms of artificial
intelligence, this implies that research lines joining artificial
life and artificial intelligence that may address a form of {}``somatic
computation'' (soma - body) may be more effective in addressing cognition
in biological systems (including human systems), a point that is addressed
in \cite{key-45}. 

In this sense, the main value of QuANNs is mostly as mathematical
and physical models for network-based quantum computation, both QuANNs
as well as ANNS can be approached mathematically as general models
of computing networks, a point, therefore, remains, as explored in
the present work, once we have network-based quantum computation and
examples of biological systems exploring quantum effects, we are led
to quantum networked computation, and, in this case, the framework
of QuANNs may provide for a tool to explore such cases independently
of what the nodes represent (actual neurons or any other active quantum
processing). Thus, there is high value of QuANNs as conceptual tools
for both applied research on quantum (bio)technology using network-based
quantum computation%
\footnote{We are considering here both hybrid technologies (with incorporated
biological components) as well as biologically-inspired technologies.
For instance, the research on the role of quantum mechanics in photosynthesis
may lead to more efficient solar energy-based technologies. Also,
in a more advanced state of quantum technologies, the possibility
of implementing evolutionary quantum computation based on QuANNs may
lead to more adaptive artificial intelligent systems with higher computing
power and smaller processing times, yielding faster and more efficient
responses (quantum-based agility solutions for intelligent systems).%
}, as well as for fundamental research in quantum biology and complex
quantum systems science.


\begin{thebibliography}{References}
\bibitem[1]{key-1}Lafontaine C (2004) L'Empire cybernétique : Des
machines à penser à la pensée machine (\emph{The cybernetic empire:
From the thinking machines to the machine thinking}). Seuil, France.

\bibitem[2]{key-2}Wiener N (1948) Cybernetics: Or Control and Communication
in the Animal and the Machine. Hermann \& Cie, Paris.

\bibitem[3]{key-3}Turing A (1936) On Computable Numbers, with an
Application to the Entscheidungsproblem. Proceedings of the London
Mathematical Society, Series 2, 42:230 - 265.

\bibitem[4]{key-4}Turing, A (1950) Computing Machinery and Intelligence.
\emph{Mind} LIX 236:433-460.

\bibitem[5]{key-5}Von Neumann J (1963) The General and Logical Theory
of Automata. In: Taub AH (ed.) John von Neumann - Collected Works,
Design of Computers, Theory of Automata and Numerical Analysis, Vol.5,
Pergamon Press, Macmillan, New York.

\bibitem[6]{key-6}Langton C (ed.) (1995) Artificial Life: An Overview.
MIT, Cambridge.

\bibitem[7]{key-7}Everett H (1955) The Theory of the Universal Wavefunction.
PhD Manuscript, In: DeWitt R and Graham N (eds.) (1973) The Many-Worlds
Interpretation of Quantum Mechanics. Princeton Series in Physics,
Princeton University Press.

\bibitem[8]{key-8}Everett H (1957) 'Relative state' formulation of
quantum mechanics. Rev. of Mod. Physics 29 (3):454\textendash{}462.

\bibitem[9]{key-9}Wiesner S (1983) Conjugate Coding. ACM SIGACT News
Vol. 15, Issue 1, Winter-Spring:78-88.

\bibitem[10]{key-10}Meyer DA (1999) Quantum strategies. Phys. Rev.
Lett. 82:1052-1055.

\bibitem[11]{key-11}Piotrowski EW and S\l{}adkowski J (2003) An invitation
to quantum game theory. Int. Journ. of Theor. Phys. 42 (5):1089\textendash{}1099.

\bibitem[12]{key-12}Holevo AS (1973) Bounds for the quantity of information
transmitted by a quantum communication channel. Problems of Information
Transmission, 9:177-183, 1973.

\bibitem[13]{key-13}Feynman, R (1982) Simulating physics with computers.
Int. Journ. of Theor. Phys. 21:467-488, 1982.

\bibitem[14]{key-14}Benioff, P (1982) Quantum mechanical hamiltonian
models of Turing machines. Journ. of Stat. Phys. 29 (3): 515\textendash{}546.

\bibitem[15]{key-15}Deutsch D (1985) Quantum theory, the Church-Turing
Principle and the universal quantum computer. Proc. R. Soc. Lond.
A:400-497.

\bibitem[16]{key-16}Nielsen MA and Chuang IL (2010) Quantum Computation
and Quantum Information. Cambridge University Press, Cambridge.

\bibitem[17]{key-17}Löwdin P-O (1963) Proton tunneling in DNA and
its biological implications. Rev. of Mod. Phys., 35 (3):724-732.

\bibitem[18]{key-18}McFadden J and Al-Khalili J (1999) A quantum
mechanical model of adaptive mutation. BioSystems, 50:203-211.

\bibitem[19]{key-19}MacFadden J (2000) Quantum Evolution. Harper
Collins, United Kingdom.

\bibitem[20]{key-20}Ball P (2011) Physics of life: The dawn of quantum
biology. Nature, 474:272-274.

\bibitem[21]{key-21}Lloyd S (1988) Black Holes, Demons and the Loss
of Coherence. PhD Thesis, The Rockefeller University.

\bibitem[22]{key-22}McCulloch W and Pitts W (1943) A logical calculus
of the ideas immanent in nervous activity. Bulletin of Mathematical
Biophysics, 7:115 - 133.

\bibitem[23]{key-23}Gonçalves CP (2012) Financial Turbulence, Business
Cycles and Intrinsic Time in an Artificial Economy, Algorithmic Finance
(2012), 1:2:141-156 

\bibitem[24]{key-24}Gonçalves CP (2013) Quantum Financial Economics
Risk and Returns. Journal of Systems Science and Complexity, April
2013, Volume 26, Issue 2:187-200.

\bibitem[25]{key-25}Rosenblatt F (1957), The Perceptron - a perceiving
and recognizing automaton. Report 85-460-1, Cornell Aeronautical Laboratory.

\bibitem[26]{key-26}Rosenblatt F (1958) The perceptron: A probabilistic
model for information storage and organization in the brain. Psychological
Review, Vol 65(6):386-408.

\bibitem[27]{key-27}McCulloch W and Pitts W (1943) A logical calculus
of the ideas immanent in nervous activity. Bulletin of Mathematical
Biophysics, 7:115 - 133.

\bibitem[28]{key-28}Dupuy J-P (2000) The Mechanization of the Mind.
Translation by MB DeBevoise, Princeton University Press, Princeton.

\bibitem[29]{key-29}Müller B, Reinhardt J and Strickland MT (1995)
Neural Networks An Introduction. Springer-Verlag, Berlin.

\bibitem[30]{key-30}Asaro P (2007) Heinz von Foerster and the Bio-Computing
Movements of the 1960s. In: Müller A and Müller KH (Eds) An Unfinished
Revolution? Heinz von Foerster and the Biological Computer Laboratory,
Edition Echoarum, Vienna.

\bibitem[31]{key-31}Eccles JC (1994) How the Self Controls its Brain.
Springer-Verlag, Berlin.

\bibitem[32]{key-32}Beck F and Eccles JC (1998) Quantum processes
in the brain: A scientific basis of consciousness. Cognitive Studies:
Bulletin of the Japanese Cognitive Science Society 5 (2): 95\textendash{}109.

\bibitem[33]{key-33}Penrose R and Hameroff S (2011) Consciousness
in the Universe: Neuroscience, Quantum Space-Time Geometry and Orch
OR Theory. Journal of Cosmology, Vol. 14, http://journalofcosmology.com/Consciousness160.html. 

\bibitem[34]{key-34}Tegmark M (2000) Importance of quantum decoherence
in brain processes. Phys. Rev. E 61 (4):4194-4206.

\bibitem[35]{key-35}Chrisley R (1995) Quantum learning. In Pylkkänen
P and Pylkkö P (eds.), New directions in cognitive science: Proceedings
of the international symposium, Saariselka, 4-9 August 1995, Lapland,
Finland, Finnish Artificial Intelligence Society, Helsinki.

\bibitem[36]{key-36}Kak S (1995) Quantum Neural Computing. Advances
in Imaging and Electron Physics, vol. 94:259-313.

\bibitem[37]{key-37} Menneer T and Narayanan A (1995) Quantum-inspired
Neural Networks. technical report R329, Department of Computer Science,
University of Exeter, Exeter, United Kingdom.

\bibitem[38]{key-38}Behrman EC, Niemel J, Steck JE, Skinner SR (1996)
A Quantum Dot Neural Network. In: Toffoli T and Biafore M, Proceedings
of the 4th Workshop on Physics of Computation, Elsevier Science Publishers,
Amsterdam.

\bibitem[39]{key-39}Menneer T (1998) Quantum Artificial Neural Networks.
Ph. D. thesis, The University of Exeter, UK.

\bibitem[40]{key-40}Ivancevic VG and Ivancevic TT (2010) Quantum
Neural Computation. Springer, Dordrecht.

\bibitem[41]{key-41}Zizzi P (2003) Emergent Consciousness: From the
Early Universe to Our Mind. NeuroQuantology,Vol.3:295-311.

\bibitem[42]{key-42}Zizzi P (2005) A Minimal Model for Quantum Gravity.
Mod.Phys.Lett. A20:645-653.

\bibitem[43]{key-43}Damásio, A (1994) Descartes' Error: Emotion,
Reason, and the Human Brain. HarperCollins Publishers, New York.

\bibitem[44]{key-44}Damásio, A (1999) The Feeling of What Happens:
Body and Emotion in the Making of Consciousness- Harcourt, San Diego.

\bibitem[45]{key-45}Gonçalves, CP (2014) Emotional Responses in Artificial
Agent-Based Systems: Reflexivity and Adaptation in Artificial Life.
arXiv:1401.2121v1 {[}cs.AI{]} (http://arxiv-web3.library.cornell.edu/abs/1401.2121).\end{thebibliography}
\end{document}